\documentclass[conference]{IEEEtran}
\IEEEoverridecommandlockouts
\usepackage{cite}
\usepackage{amsmath,amssymb,amsfonts}
\usepackage{algorithmic}
\usepackage{graphicx}
\usepackage{stfloats}
\usepackage{textcomp}
\usepackage{xcolor}
\usepackage{subfig}
\def\BibTeX{{\rm B\kern-. 05em{\sc i\kern-. 025em b}\kern-. 08em
    T\kern-. 1667em\lower. 7ex\hbox{E}\kern-. 125emX}}

\usepackage[normalem]{ulem}
\usepackage{rotating}
\usepackage{tabularray}
\usepackage{hyperref}

\begin{document}

\title{2DXformer: Dual Transformers for Wind Power Forecasting with Dual Exogenous Variables
}

\author{
    \IEEEauthorblockN{Yajuan Zhang\textsuperscript{1,3}, Jiahai Jiang\textsuperscript{1}, Yule Yan\textsuperscript{1}, Liang Yang\textsuperscript{1,3}, Ping Zhang\textsuperscript{1,2}}
    \IEEEauthorblockA{
        \textit{\textsuperscript{1}School of Artificial Intelligence, Hebei University of Technology, Tianjin, China} \\
        \textit{\textsuperscript{2}State Key Laboratory of Reliability and Intelligence of Electrical Equipment, Hebei University of Technology, Tianjin, China}\\
        \textit{\textsuperscript{3}Hebei Province Key Laboratory of Big Data Calculation, Hebei University of Technology, Tianjin, China}\\
        \{zhangyajuan, zhangping\}@hebut.edu.cn, \{202232805044, 202222802015\}@stu.hebut.edu.cn, yangliang@vip.qq.com
    }
}


\maketitle

\begin{abstract}
Accurate wind power forecasting can help formulate scientific dispatch plans, which is of great significance for maintaining the safety, stability, and efficient operation of the power system.  In recent years, wind power forecasting methods based on deep learning have focused on extracting the spatiotemporal correlations among data, achieving significant improvements in forecasting accuracy.  However, they exhibit two limitations.  First, there is a lack of modeling for the inter-variable relationships, which limits the accuracy of the forecasts.  Second, by treating endogenous and exogenous variables equally, it leads to unnecessary interactions between the endogenous and exogenous variables, increasing the complexity of the model.  In this paper, we propose the 2DXformer, which, building upon the previous work's focus on spatiotemporal correlations, addresses the aforementioned two limitations.  Specifically, we classify the inputs of the model into three types: exogenous static variables, exogenous dynamic variables, and endogenous  variables.  First, we embed these variables as variable tokens in a channel-independent manner.  Then, we use the attention mechanism to capture the correlations among exogenous variables.  Finally, we employ a multi-layer perceptron with residual connections to model the impact of exogenous variables on endogenous  variables.  Experimental results on two real-world large-scale datasets indicate that our proposed 2DXformer can further improve the performance of wind power forecasting.  The code is available in this repository: \href{https://github.com/jseaj/2DXformer}{https://github.com/jseaj/2DXformer}. 

\end{abstract}

\begin{IEEEkeywords}
wind power forecasting, spatiotemporal forecasting, exogenous variables, variable correlation
\end{IEEEkeywords}

\section{Introduction}
The ongoing enhancement of living standards and societal development has led to a persistent rise in human energy demand. Traditional fossil fuels (coal, oil, natural gas) cause pollution, environmental damage, and global warming \cite{wang2019approaches}. In contrast, wind energy is a pollution-free, widely distributed renewable source, offering inexhaustibility, sustainability, and reasonable pricing, thus attracting global attention \cite{liu2019data}. However, its randomness and intermittency pose challenges for grid load balancing and power dispatching \cite{lai2023dual}. Accurate wind power forecasting can help develop scientific dispatching plans, mitigating these challenges and ensuring the power system's long-term stability and reliability \cite{liu2020new}.

In recent years, deep learning methods have gained significant attention in wind power forecasting due to their strong capability in modeling complex nonlinear relationships. Current deep learning-based models typically involve multiple turbines within a wind farm and address the task as a spatio-temporal prediction problem rather than merely a time series forecasting task.
Over the years, Researchers have developed various methods to extract spatio-temporal correlations among wind turbines. Li et al. \cite{li2022deep} used the k-nearest neighbors algorithm to identify neighboring nodes of the target turbine, augmenting its data to capture spatial correlations and employing GRU to extract temporal correlations. Yu et al. \cite{yu2019scene} mapped turbine data onto a plane to form a state graph, generating multi-channel images based on relative positions and using CNNs to capture spatio-temporal correlations. Liao et al. \cite{liao2022short} presented a power prediction method using graph neural networks, constructing the graph's adjacency matrix based on Pearson correlation to represent spatial topology, and combining graph convolutional networks with LSTM to capture intricate spatio-temporal features. Zhang et al. introduced the HSTTN model \cite{ijcai2023p700}, which pioneered the use of transformer architecture in wind power forecasting, leveraging its ability to capture spatial correlations and long-term temporal dependencies, showing outstanding performance in long-term wind power prediction.

Despite significant progress in wind power forecasting by previous researchers, two key limitations persist.  These limitations, which stem from a lack of extensive and thorough exploration in critical areas, still offer opportunities for enhancing the effectiveness of wind power forecasting. 

\textit{Limitation 1:} Lack of modeling for inter-variable relationships.  Wind power data is a multi-channel signal, with each turbine recording several variables such as power and wind speed (the number and nature of these variables depend on the specific dataset).  However, existing work on wind power forecasting often treats all variables as a whole for modeling.  Specifically, they map all variables at each time point to the latent space as multi-dimensional features. This indiscriminate approach can result in excessive smoothing, thereby compromising the prediction accuracy.

\textit{Limitation 2: }Treating exogenous and endogenous variables equally. Inspired by \cite{wang2024timexer}, we define wind power as the endogenous variable and other recorded variables (e.g., wind speed, temperature) as exogenous. Relying solely on endogenous variables is insufficient for accurate wind power forecasting due to the critical role of exogenous variables. Existing studies often merge these variables indiscriminately as model inputs, complicating the model and potentially reducing prediction accuracy due to unnecessary interactions.

To address the aforementioned two limitations, we propose the 2DXformer: a \underline{D}ual Transformer with \underline{D}ual e\underline{X}ogenous variables.  The Dual exogenous variables refer to the classification of exogenous variables into two categories: exogenous dynamic variables, which are time-based embeddings derived from timestamps, and exogenous static variables, such as wind speed and temperature, which are system records. In the 2DXformer, we first encode variables into feature representations using the embedding method from ITransformer\cite{liu2024itransformer}. We then use two separate Transformer blocks: EnTBlock for endogenous variables and ExTBlock for exogenous variables. The output from ExTBlock is processed through a ResidualMLP to assess the influence of exogenous variables on the endogenous ones. This approach mitigates \textit{limitation 2} by treating these variables distinctly. Each block employs an attention mechanism to capture spatial correlations; ExTBlock focuses on correlations among exogenous variables, while EnTBlock uses a ResidualMLP to address correlations between both types. This dual consideration successfully tackles \textit{limitation 1}. In summary, the contributions of this paper include:
\begin{itemize}
    \item We propose 2DXformer, an innovative transformer-based model for wind power forecasting, that can also be applied to other multivariate spatiotemporal sequence forecasting tasks.  
    \item We propose an innovative approach to modeling exogenous variables.  This method not only accounts for the correlations between variables but also circumvents unnecessary interactions between endogenous and exogenous variables. 
    \item We conducted extensive experiments on datasets collected from two real wind farms, showcasing the effectiveness of our method via comparative analyses and ablation studies.  
\end{itemize}

\section{Related work} \label{sec2}

\subsection{Forecasting with exogenous variables}
In statistical methods, researchers have acknowledged the significance of exogenous variables, extending the classical ARIMA model \cite{bartholomew1971time} to ARIMAX \cite{williams2001multivariate} and SARIMAX \cite{vagropoulos2016comparison} to incorporate exogenous variables and bolster the model's predictive prowess.  While including more exogenous variables can enrich predictions, it may also create unnecessary interactions, potentially reducing accuracy.  In the practice of wind power forecasting, evidence suggests that the introduction of too many exogenous variables can diminish accuracy and prolong training time \cite{amjady2011wind}.  An effective and prevalent approach involves employing feature selection algorithms like Correlation Analysis (CA) \cite{liang2016short} and Principal Component Analysis (PCA) \cite{kong2015wind} to sift through the exogenous variables.  These studies acknowledge that the influence of exogenous variables on endogenous variables is not invariably advantageous.  However, their focus on the selection of exogenous variables overlooks the need for a mechanism to differentiate between endogenous and exogenous variables. Recently, TiDE\cite{das2023longterm} and TimeXer\cite{wang2024timexer} have further investigated modeling schemes for exogenous variables, implementing approaches that distinguish between the modeling of endogenous and exogenous variables.  

This paper extends these works by: (1) further classifying exogenous variables into static and dynamic categories, which is more suitable for wind power forecasting, and (2) investigating the inter-correlations among exogenous variables more comprehensively.

\section{Preliminaries} \label{sec3}

In this paper, the forecasting scenario is situated in a wind farm with $N$ wind turbines.  The objective of wind power prediction is to leverage historical wind power data $X = \{ X_{t}^{(1)}, X_{t}^{(2)}, \cdots, X_{t}^{(N)}\}_{t = 1}^{H} \in \mathbb{R}^{H \times N \times 1}$ and historical exogenous variables $Z^{s} = \{ Z_{t}^{s,(1)}, Z_{t}^{s,(2)}, \cdots, Z_{t}^{s,(N)} \}_{t = 1}^{H} \in \mathbb{R}^{H \times N \times C}$, such as wind speed and direction, to forecast the wind power output $\widehat{Y} = \{ \widehat{Y}_{t}^{(1)}, \widehat{Y}_{t}^{(2)} , \cdots, \widehat{Y}_{t}^{(N)}\}_{t = H + 1}^{H + P} \in \mathbb{R}^{P \times N \times 1}$ over the next $P$ steps.  Here, $X_{t}^{(i)} \in \mathbb{R}$ represents the power output recorded by turbine $i$ at time $t$, $Z_{t}^{s,(i)} \in \mathbb{R}^{C}$ denotes the exogenous static variables such as wind speed and direction for turbine $i$ at time $t$, and $\widehat{Y}_{t}^{(i)} \in \mathbb{R}$ stands for the predicted value for turbine $i$ at time $t$; $C$ indicates the number of exogenous static variables, $H$ represents the lookback length, and $P$ signifies the forecast horizon. 
Mathematically, the prediction of wind power can be described by the following expression:
\begin{equation}
    {{\widehat{Y}}}=f\left( {{X}},Z^{s},Z_{{}}^{d};\theta  \right)
\end{equation}
where $f\left( \cdot ;\theta  \right)$ denotes the prediction model parameterized by $\theta$, and ${{Z}^{d}} \in {{\mathbb{R}}^{{H}'\times N\times {{C}_{d}}}}$ corresponds to the exogenous dynamic variables, with ${{C}_{d}}$ indicating the number of  exogenous dynamic variables.

\section{The Proposed Method} \label{sec4}
\begin{figure}[!t]
\centering
\includegraphics[width=0.49\textwidth]{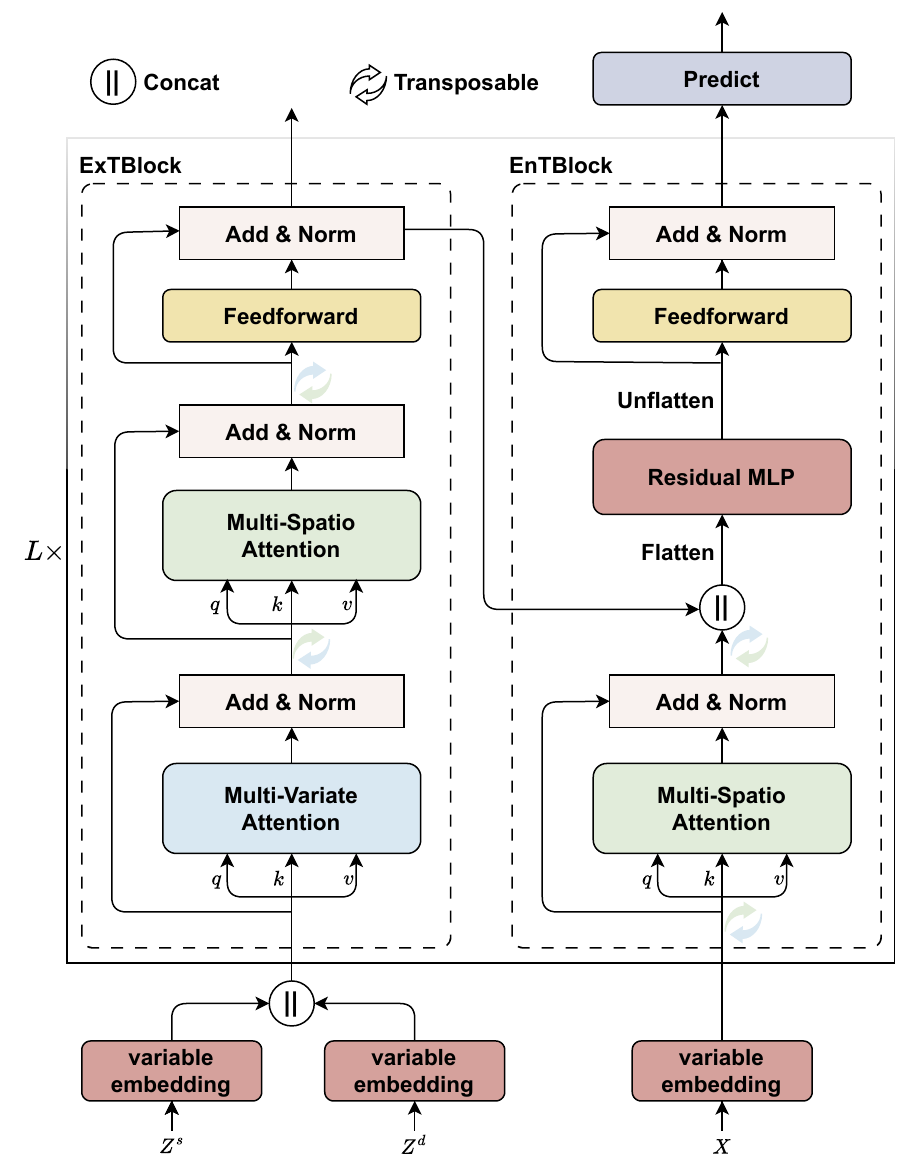}
\caption{Overall structure of the proposed model. }
\label{fig1}
\end{figure}
As depicted in Figure~\ref{fig1}, our proposed 2DXformer model is based on an Encoder-Only Transformer architecture, comprising three types of variable embeddings from the bottom up and $L$ layers, with the multi-step prediction task being executed by a linear layer.  Each layer is composed of ExTBlock and EnTBlock.  

In this section, we will detail the model by discussing variable embeddings, ExTBlock, and EnTBlock respectively, and then introduce the prediction and training process.

\subsection{Variable embedding} \label{sec4. 2}

In this paper, we adopt the method consistent with the literature \cite{liu2024itransformer}, resulting in endogenous variable embedding ${{V}_{en}} \in {{\mathbb{R}}^{N \times 1 \times D}}$, exogenous static variable embedding ${{V}_{sex}} \in {{\mathbb{R}}^{N \times C \times D}}$, and exogenous dynamic variable embedding ${{V}_{dex}} \in {{\mathbb{R}}^{N \times {{C}_{d}} \times D}}$. The above process can be formalized as:
\begin{equation}
    {{V}_{en}}=EnVarEmb\left( {{X}} \right)
\end{equation}
\begin{equation}
    V_{sex}^{(i)}=SExVarEmb\left( Z^{s,(i)} \right),i\in \left\{ 1,2,\cdots ,C \right\}
\end{equation}
\begin{equation}
    V_{dex}^{(i)}=DExVarEmb\left( Z_{{}}^{d,(i)} \right),i\in \left\{ 1,2,\cdots ,{{C}_{d}} \right\}
\end{equation} 
Here, $EnVarEmb\left( \cdot  \right)$, $SExVarEmb\left( \cdot  \right)$, and $DExVarEmb\left( \cdot  \right)$ represent embedding methods for different variables.  Structurally, they share a common architecture: a Multi-Layer Perceptron with residuals (ResidualMLP). 

It’s important to note that the exogenous dynamic variable ${{Z}^{d}}$ is generated indirectly from historical timestamps rather than recorded data.   Specifically, We extract three types of temporal information: the sample's position within the day, the month of collection, and the day of the year.  This information is encoded using three learnable embedding matrices: a diurnal embedding matrix ${{E}_{T}} \in {{\mathbb{R}}^{{{N}_{t}} \times {{C}_{e}}}}$, a monthly embedding matrix ${{E}_{M}} \in {{\mathbb{R}}^{12 \times {{C}_{e}}}}$, and an yearly embedding matrix ${{E}_{Y}} \in {{\mathbb{R}}^{366 \times {{C}_{e}}}}$, here ${{C}_{e}}$ denoting the embedding dimension. The resulting diurnal, monthly, and yearly embedding features ${{F}_{T}}$, ${{F}_{M}}$, and ${{F}_{Y}}$ are concatenated to form ${{Z}^{d} \in \mathbb{R}^{H' \times N \times C_{d}}}$, where it is evident that $C_{d} = 3 \times C_{e}$. The above
process can be formalized as:
\begin{equation}
    {{Z}^{d}}={{F}_{T}}||{{F}_{M}}||{{F}_{Y}}
\end{equation}	 
Since the aforementioned temporal information after the moment $t$ is available, consequently, the sequence length $H'$ of ${{Z}^{d}}$ may differ from the sequence length $H$ of endogenous variables and exogenous static variables. 

\subsection{ExTBlock}
ExTBlock is a transformer block specifically designed for processing exogenous variables, aimed at capturing the inter-variable relationships and spatial correlations of these exogenous variables.  These objectives are addressed by two built-in multi-head attention mechanisms within the ExTBlock.  The implementation details are as follows:

First, concatenate ${{V}_{dex}}$ and ${{V}_{sex}}$ to form the embedding representation of the exogenous variables, ${{V}_{ex}} \in {{\mathbb{R}}^{N \times (C + {{C}_{d}}) \times D}}$, which serves as the initial input to ExTBlock, denoted as $H_{ex}^{(0)} = {{V}_{ex}}$:
\begin{equation}
    {{V}_{ex}}={{V}_{dex}}||{{V}_{sex}}
\end{equation}
In the $l_{th}$ layer of the ExTBlock, the output from the $(l-1)_{th}$ layer, $H_{ex}^{(l-1)} \in {{\mathbb{R}}^{N \times (C + {{C}_{d}}) \times D}}$, is initially processed by the first multi-head self-attention module to capture the inter-variable relationships.  The derived hidden features are denoted as $H_{v,ex}^{(l)} \in {{\mathbb{R}}^{N \times (C + {{C}_{d}}) \times D}}$.  The computation is as follows:
\begin{equation}
    H_{v,ex}^{(l)}=LN\left( H_{ex}^{(l-1)}+SelfAttn\left( H_{ex}^{(l-1)} \right) \right)
\end{equation}
Here, $LN(\cdot)$ represents layer normalization, and $SelfAttn(\cdot)$ indicates the multi-head self-attention mechanism.  To utilize the second multi-head self-attention mechanism for capturing spatial correlations, it is necessary to perform a transpose operation on the first two dimensions of the input $H_{v,ex}^{(l)} \in {{\mathbb{R}}^{N \times (C + {{C}_{d}}) \times D}}$, resulting in $H_{v,ex}^{(l)}$ being transformed into $H_{v,ex}^{(l)} \in {{\mathbb{R}}^{(C + {{C}_{d}}) \times N \times D}}$.  Subsequently, the output $H_{s,ex}^{(l)} \in {{\mathbb{R}}^{(C + {{C}_{d}}) \times N \times D}}$ is derived in a similar manner as before:
\begin{equation}
    H_{s,ex}^{(l)}=LN\left( H_{v,ex}^{(l)}+SelfAttn\left( H_{v,ex}^{(l)} \right) \right)
\end{equation}
After performing the transpose operation on $H_{s,ex}^{(l)}$, it is transformed into $H_{s,ex}^{(l)} \in {{\mathbb{R}}^{N \times (C + {{C}_{d}}) \times D}}$.  Subsequently, it is processed by a FeedForward network to derive the final output of the $l_{th}$ layer ExTBlock, denoted as $H_{ex}^{(l)} \in {{\mathbb{R}}^{N \times (C + {{C}_{d}}) \times D}}$. 

\subsection{EnTBlock}
The EnTBlock, tailored to process endogenous variables, mirrors the ExTBlock in its application of multi-head self-attention to capture spatial correlations among endogenous variables.  However, when modeling the influence of exogenous variables on endogenous variables, we opt for a simple ResidualMLP over multi-head attention, a choice validated by ablation studies detailed in Section \ref{sec:ab}.  The specifics of the EnTBlock are as follows:

We consider the embeddings of endogenous variables ${{V}_{en}}$ as the initial input to the EnTBlock, denoted as $H_{en}^{(0)} = {{V}_{en}}$.  In the $l_{th}$ layer of the EnTBlock, spatial relationships are initially extracted via multi-head self-attention, followed by the modeling of exogenous variables' influence on endogenous variables through ResidualMLP.  Analogous to the ExTBlock, the input $H_{en}^{(l-1)} \in {{\mathbb{R}}^{N \times 1 \times D}}$ is transposed to $H_{en}^{(l-1)} \in {{\mathbb{R}}^{1 \times N \times D}}$, facilitating the extraction of spatial relationships through multi-head self-attention:
\begin{equation}
    H_{s,en}^{(l)}=LN\left( H_{en}^{(l - 1)}+SelfAttn\left( H_{en}^{(l - 1)} \right) \right)
\end{equation}
The output $H_{s,en}^{(l)} \in {{\mathbb{R}}^{1 \times N \times D}}$ is transformed back to $H_{s,en}^{(l)} \in {{\mathbb{R}}^{N \times 1 \times D}}$.  Prior to capturing the influence of exogenous variables on endogenous variables, it is essential to reduce the dimensions of $H_{s,en}^{(l)}$ and $H_{ex}^{(l)}$ to $H_{s,en}^{(l)} \in {{\mathbb{R}}^{N \times (1 \times D)}}$ and $H_{ex}^{(l)} \in {{\mathbb{R}}^{N \times ((C + {{C}_{d}}) \times N)}}$, respectively, before feeding them into the ResidualMLP:
\begin{equation}
    H_{v,en}^{(l)}=ResidualMLP\left( H_{s,en}^{(l)}||H_{ex}^{(l)} \right)
\end{equation}
Where $H_{v,en}^{(l)} \in {{\mathbb{R}}^{N \times (1 \times D)}}$.  After transforming the dimension of $H_{v,en}^{(l)}$ back to $H_{v,en}^{(l)} \in {{\mathbb{R}}^{N \times 1 \times D}}$, it is subsequently processed by a FeedForward network to derive the final output of the $l^{th}$ layer EnTBlock, denoted as $H_{en}^{(l)} \in {{\mathbb{R}}^{N \times 1 \times D}}$. 

\subsection{Prediction and Training}
In 2DXformer, the final $P$-step prediction is fully implemented by the MLP, specifically,
\begin{equation}
    {\widehat{Y}}=MLP\left( H_{en}^{\left( L \right)} \right)
\end{equation} 
Where $\widehat{Y} \in \mathbb{R}^{P \times N \times 1}$ denotes the predicted values for the future $P$ steps, while ${H_{en}^{(L)}}$ indicates the output from the last layer of the EnTBlock.  To assess the discrepancy between the predicted values and the ground truth, we use the smooth L1\cite{girshick2015fast} as the loss function.

\section{Experiments} \label{sec5}
\subsection{ Experimental Setup}
\subsubsection{Datasets}
We perform experiments with the datasets SDWPF and HHL16, which are obtained from two actual wind farms:
\begin{itemize}
    \item SDWPF\cite{zhou2022sdwpf}: This dataset is sourced from a wind farm with 134 turbines, collected by Longyuan Power Group Corporation.  The SDWPF dataset samples every 10 minutes, encompassing a total of 4,727,520 data records over 245 days.  Each record includes 13 features, 
    such as sampling time, wind speed, temperature, yaw angle, power, etc.
    \item HHL16: The data is extracted from the SCADA system of a power station located in Hebei Province, China, consisting of 33 wind turbines.  The dataset samples every 10 minutes, encompassing a total of 1,739,232 data records from January 1, 2016, to December 31, 2016.  Each sample includes 9 features, such as sampling time, wind speed, temperature, power, and others. 
\end{itemize}

In this paper, we define the endogenous variable as solely the wind power output, while all other variables, excluding power, timestamp, and wind turbine number, are considered exogenous static variables.  Consequently, the number of exogenous static variables in SDWPF dataset and HHL16 dataset is 9 and 6 respectively.

We divide all datasets into training, validation, and testing sets according to the 7:2:1 ratio.  
Additionally, for missing values in the dataset, we employ a combination of forward-filling and backward-filling for interpolation.  Lastly, we normalize the entire dataset, excluding the labels, using the Z-score method and apply de-normalization to the model's output.

\subsubsection{Baselines and Evaluation Metrics}
In this paper, we evaluate our proposed 2DXformer against eight distinct deep learning models.  These comparison models are classified into the following three categories:
\begin{itemize}
    \item Traditional Deep Learning Models (Trad): MLP, GRU\cite{cho2014learning}, and Transformer\cite{vaswani2017attention}.   
    
    \item Spatio Temporal Based Model (ST-B): 
    AGCRN\cite{bai2020adaptive}, MegaCRN\cite{jiang2023spatio}, and STAEformer\cite{liu2023spatio}.

    \item Variable Base Model (V-B): TiDE\cite{das2023longterm} and TimeXer\cite{wang2024timexer}. 
\end{itemize}

All methods were conducted on two preprocessed datasets under identical experimental conditions to ensure a fair comparison. 
For evaluation, two commonly employed metrics, Mean Absolute Error (MAE) and Root Mean Square Error (RMSE), are utilized to compare the predicted values $\widehat{Y}$ against the ground truth $Y$.

\subsubsection{Implementation Details}
All experiments are conducted on an NVIDIA GeForce RTX 4090 GPU with Pytorch 2.1.0.  Some important hyperparameter settings of 2DXformer are as follows:
$D = 64$, $C_e = 16$, $C_d = 48$ and $L = 3$.
During model training, we employed the Adam optimizer, initiating the learning rate at $5 \times 10^{-4}$ and gradually diminishing it throughout the training process.

\subsection{Performance Comparison}

\begin{table*}
\centering
\caption{The performance of different methods on two datasets.  The units for the two evaluation metrics, MAE and RMSE, are both in kW.  The input length for all baselines is set to 36, and `PL=12 (2h)' indicates a prediction length of 12. \textbf{Bold} signifies the best result, and \uline{underline} signifies the second best result.}
\label{t1}
\begin{tblr}{
  row{2} = {c},
  row{3} = {c},
  row{12} = {c},
  cell{1}{1} = {r=3}{c},
  cell{1}{2} = {r=3}{},
  cell{1}{3} = {c=3}{c},
  cell{1}{6} = {c=3}{c},
  cell{4}{1} = {r=3}{c},
  cell{4}{3} = {c},
  cell{4}{4} = {c},
  cell{4}{5} = {c},
  cell{4}{6} = {c},
  cell{4}{7} = {c},
  cell{4}{8} = {c},
  cell{5}{3} = {c},
  cell{5}{4} = {c},
  cell{5}{5} = {c},
  cell{5}{6} = {c},
  cell{5}{7} = {c},
  cell{5}{8} = {c},
  cell{6}{3} = {c},
  cell{6}{4} = {c},
  cell{6}{5} = {c},
  cell{6}{6} = {c},
  cell{6}{7} = {c},
  cell{6}{8} = {c},
  cell{7}{1} = {r=3}{c},
  cell{7}{3} = {c},
  cell{7}{4} = {c},
  cell{7}{5} = {c},
  cell{7}{6} = {c},
  cell{7}{7} = {c},
  cell{7}{8} = {c},
  cell{8}{3} = {c},
  cell{8}{4} = {c},
  cell{8}{5} = {c},
  cell{8}{6} = {c},
  cell{8}{7} = {c},
  cell{8}{8} = {c},
  cell{9}{3} = {c},
  cell{9}{4} = {c},
  cell{9}{5} = {c},
  cell{9}{6} = {c},
  cell{9}{7} = {c},
  cell{9}{8} = {c},
  cell{10}{1} = {r=2}{c},
  cell{10}{3} = {c},
  cell{10}{4} = {c},
  cell{10}{5} = {c},
  cell{10}{6} = {c},
  cell{10}{7} = {c},
  cell{10}{8} = {c},
  cell{11}{3} = {c},
  cell{11}{4} = {c},
  cell{11}{5} = {c},
  cell{11}{6} = {c},
  cell{11}{7} = {c},
  cell{11}{8} = {c},
  cell{12}{1} = {c=2}{},
  vline{4} = {1}{0.05em},
  vline{6} = {1-12}{0.05em},
  hline{1,13} = {-}{0.08em},
  hline{2-3} = {3-8}{0.03em},
  hline{4,7,10,12} = {-}{0.05em},
}
Type                                      & Models      & SDWPF                &                      &                      & HHL16               &                      &                      \\
                                          &             & PL=12(2h)            & PL=24(4h)            & PL=36(6h)            & PL=12(2h)            & PL=24(4h)            & PL=36(6h)            \\
                                          &             & MAE/RMSE             & MAE/RMSE             & MAE/RMSE             & MAE/RMSE             & MAE/RMSE             & MAE/RMSE             \\
\begin{sideways}Trad\end{sideways} & MLP         & 96.51/164.57         & 114.77/190.53        & 126.70/205.19        & 133.95/250.48        & 163.92/293.37        & 185.25/324.04        \\
                                          & GRU         & 93.61/158.04         & 104.84/173.55        & 112.23/182.29        & 133.77/246.83        & 161.37/288.84        & 182.65/317.69        \\
                                          & Transformer & 78.37/135.59         & 89.95/151.48         & 97.84/162.40         & 111.02/206.97        & 129.67/234.57        & 143.09/254.53        \\
\begin{sideways}ST-B\end{sideways}    & AGCRN       & 47.34/83.96          & 53.03/93.13          & 59.09/102.80         & \uline{51.76/102.18} & 60.33/116.77         & 68.77/131.70         \\
                                          & MegaCRN     & 67.52/124.76         & 70.14/125.79         & 77.79/136.70         & 84.61/161.94         & 83.95/158.63         & 91.11/179.36         \\
                                          & STAEformer  & \uline{44.49/85.48}  & \uline{45.23/81.27}  & \textbf{44.61/79.78} & 56.92/108.73         & \uline{57.26/102.20} & \uline{53.62/100.46} \\
\begin{sideways}V-B\end{sideways}    & TimeXer     & 72.72/128.91         & 79.83/138.14         & 83.96/143.24         & 98.30/186.85         & 104.19/193.21        & 108.04/198.16        \\
                                          & TiDE        &   66.17/119.22 &  74.00/128.55 &  80.96/138.16  &    96.80/183.21   &  105.67/196.18   &   107.09/198.12   \\
2DXformer(ours)                           &             & \textbf{44.00/81.62} & \textbf{43.74/80.04} & \uline{46.28/84.31}  & \textbf{46.85/95.06} & \textbf{44.57/90.61} & \textbf{46.95/94.95} 
\end{tblr}
\end{table*}
Table~\ref{t1} presents a performance comparison between our 2DXformer method and various baseline models. 2DXformer consistently achieves near-optimal performance across all metrics and datasets. Traditional deep learning models treat wind power prediction as a time-series task, neglecting spatial modeling, which limits their effectiveness. While TiDE and TimeXer improve upon traditional models by considering inter-variable correlations, they still overlook spatial information. In contrast, spatio-temporal models like STAEformer demonstrate superior performance by incorporating both spatial and temporal relationships. Our model, by thoroughly exploring these correlations and variable interactions, excels across the two datasets.

\subsection{Ablation Studies} \label{sec:ab}
\begin{table}
\centering
\caption{Performance of different variants of 2DXformer on the SDWPF dataset.  The number of layers for all variants is set to 2. }
\label{t2}
\begin{tblr}{
  row{2} = {c},
  cell{1}{1} = {r=2}{},
  cell{1}{2} = {c},
  cell{1}{3} = {c},
  cell{1}{4} = {c},
  cell{3}{2} = {c},
  cell{3}{3} = {c},
  cell{3}{4} = {c},
  cell{4}{2} = {c},
  cell{4}{3} = {c},
  cell{4}{4} = {c},
  cell{5}{2} = {c},
  cell{5}{3} = {c},
  cell{5}{4} = {c},
  cell{6}{2} = {c},
  cell{6}{3} = {c},
  cell{6}{4} = {c},
  cell{7}{2} = {c},
  cell{7}{3} = {c},
  cell{7}{4} = {c},
  cell{8}{2} = {c},
  cell{8}{3} = {c},
  cell{8}{4} = {c},
  cell{9}{2} = {c},
  cell{9}{3} = {c},
  cell{9}{4} = {c},
  hline{1} = {1-4}{0.08em},
  hline{2} = {2-4}{0.03em},
  hline{3,9} = {-}{0.03em},
  hline{10} = {-}{0.08em},
}
Models     & PL=12(2h)            & PL=24(4h)            & PL=36(6h)            \\
           & MAE/RMSE             & MAE/RMSE             & MAE/RMSE             \\
\textit{w/o} EDV    & 59.80/109.82         & 60.58/108.05         & 62.79/110.63         \\
\textit{w/o} DEV    & \uline{48.36/87.60}  & 50.68/90.82          & 53.54/95.01          \\
\textit{w/o} ESC    & 51.46/95.80          & 53.44/95.61          & \uline{52.60/93.55}  \\
\textit{w/o} EVC    & 49.59/90.60          & \uline{50.61/90.39}  & 52.64/93.21          \\
\textit{w/o} ESVC   & 57.04/104.08         & 59.10/105.72         & 60.87/107.23         \\
Rep ByAttn & 78.08/138.35         & 80.76/140.02         & 83.29/142.29         \\
2DXformer  & \textbf{46.62/85.50} & \textbf{42.46/78.34} & \textbf{47.40/85.65} 
\end{tblr}
\end{table}
To assess the effectiveness of each component in the 2DXformer model and gauge their contributions to the overall performance, we designed six distinct variants of the 2DXformer:
\begin{itemize}
    \item Rep ByAttn: Replace the ResidualMLP in the EnTBlock with cross attention. 
    \item \textit{w/o} DEV: Without differentiate between exogenous dynamic and static variables, meaning that these variables share the same encoding block. 
    \item \textit{w/o} EDV: Without exogenous dynamic variables. 
    \item \textit{w/o} ESC: Without the attention module responsible for spatial correlations in the ExTBlock. 
    \item \textit{w/o} ESVC: Without the two attention modules responsible for spatial and variable correlations in the ExTBlock. 
    \item \textit{w/o} EVC: Without the attention module responsible for inter-variable correlations in the ExTBlock. 
\end{itemize}

We perform experiments on the SDWPF dataset using the six 2DXformer variants described above, with the results presented in Table~\ref{t2}.  Overall, each component significantly contributes to enhancing predictive performance, demonstrating their indispensability.  In terms of impact on model performance, ResidualMLP exerts the most significant influence.  Replacing ResidualMLP with attention results in a substantial decrease in the predictive capabilities of 2DXformer.  Additionally, exogenous dynamic variables are equally vital for prediction accuracy. 

\subsection{Hyperparameter Studies}
\begin{table}
\centering
\caption{Performance of different hyperparameters $L$ on two datasets. }
\label{t3}
\begin{tblr}{
  row{2} = {c},
  column{1} = {c},
  cell{1}{1} = {r=2}{},
  cell{1}{2} = {r=2}{},
  cell{1}{3} = {c},
  cell{1}{4} = {c},
  cell{1}{5} = {c},
  cell{3}{1} = {r=3}{},
  cell{3}{3} = {c},
  cell{3}{4} = {c},
  cell{3}{5} = {c},
  cell{4}{3} = {c},
  cell{4}{4} = {c},
  cell{4}{5} = {c},
  cell{5}{3} = {c},
  cell{5}{4} = {c},
  cell{5}{5} = {c},
  cell{6}{1} = {r=3}{},
  cell{6}{3} = {c},
  cell{6}{4} = {c},
  cell{6}{5} = {c},
  cell{7}{3} = {c},
  cell{7}{4} = {c},
  cell{7}{5} = {c},
  cell{8}{3} = {c},
  cell{8}{4} = {c},
  cell{8}{5} = {c},
  hline{1} = {1-5}{0.08em},
  hline{2} = {3-5}{0.03em},
  hline{3,6} = {-}{0.05em},
  hline{9} = {-}{0.08em},
}
Dataset                              & $L$ & PL=12(2h)            & PL=24(4h)            & PL=36(6h)            \\
                                     &          & MAE/RMSE             & MAE/RMSE             & MAE/RMSE             \\
\begin{sideways}SDWPF\end{sideways}  & 1        & 50.81/94.23          & 50.34/90.33          & 50.94/90.60          \\
                                     & 2        & 46.62/85.50          & \textbf{42.46/78.34} & 47.40/85.65          \\
                                     & 3        & \textbf{44.00/81.62} & 43.74/80.04          & \textbf{46.28/84.31} \\
\begin{sideways}HHL16\end{sideways} & 1        & 52.30/105.07         & 48.16/96.12          & 52.03/102.54         \\
                                     & 2        & 48.64/98.74          & 45.43/91.89          & 47.36/96.01          \\
                                     & 3        & \textbf{46.85/95.06} & \textbf{44.57/90.61} & \textbf{46.95/94.95} 
\end{tblr}
\end{table}
In this section, we investigate the effect of a crucial hyperparameter, $L$, on the 2DXformer model.  We set $L$ to $\left\{ 1,2,3 \right\}$ and conduct experiments on the SDWPF and HHL16 datasets, with the results displayed in Table~\ref{t3}.  Generally, as the number of layers increases, the model shows improvement in both MAE and RMSE metrics.  However, this trend is not entirely consistent on the SDWPF dataset.  Nonetheless, it is evident that with more layers, the time required for both training and inference also increases. 

\subsection{Visualization}
\begin{figure*}[!t]
\centering
\subfloat[]{\includegraphics[width=3.2in]{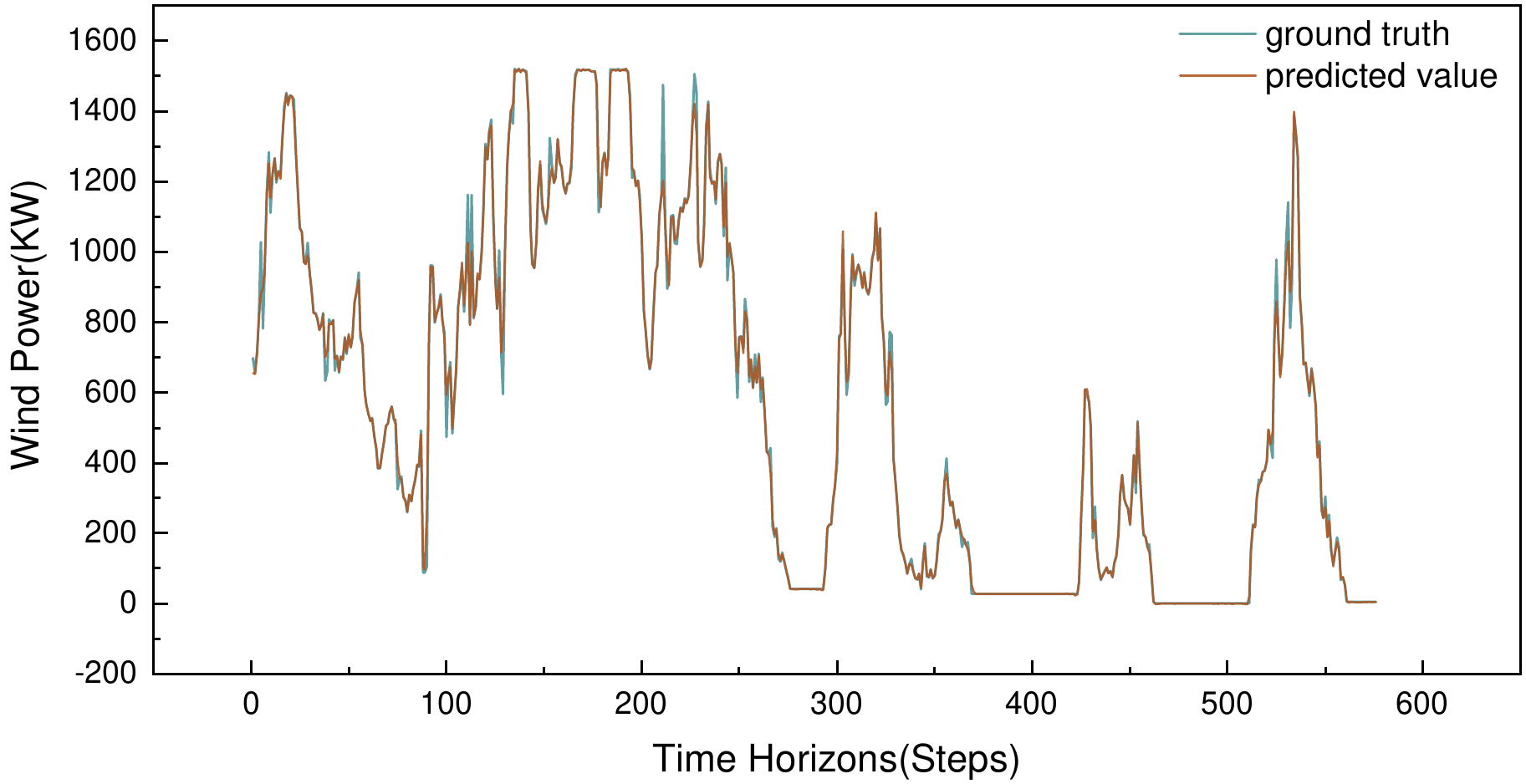}%
\label{fig3a}}
\hfil
\subfloat[]{\includegraphics[width=3.2in]{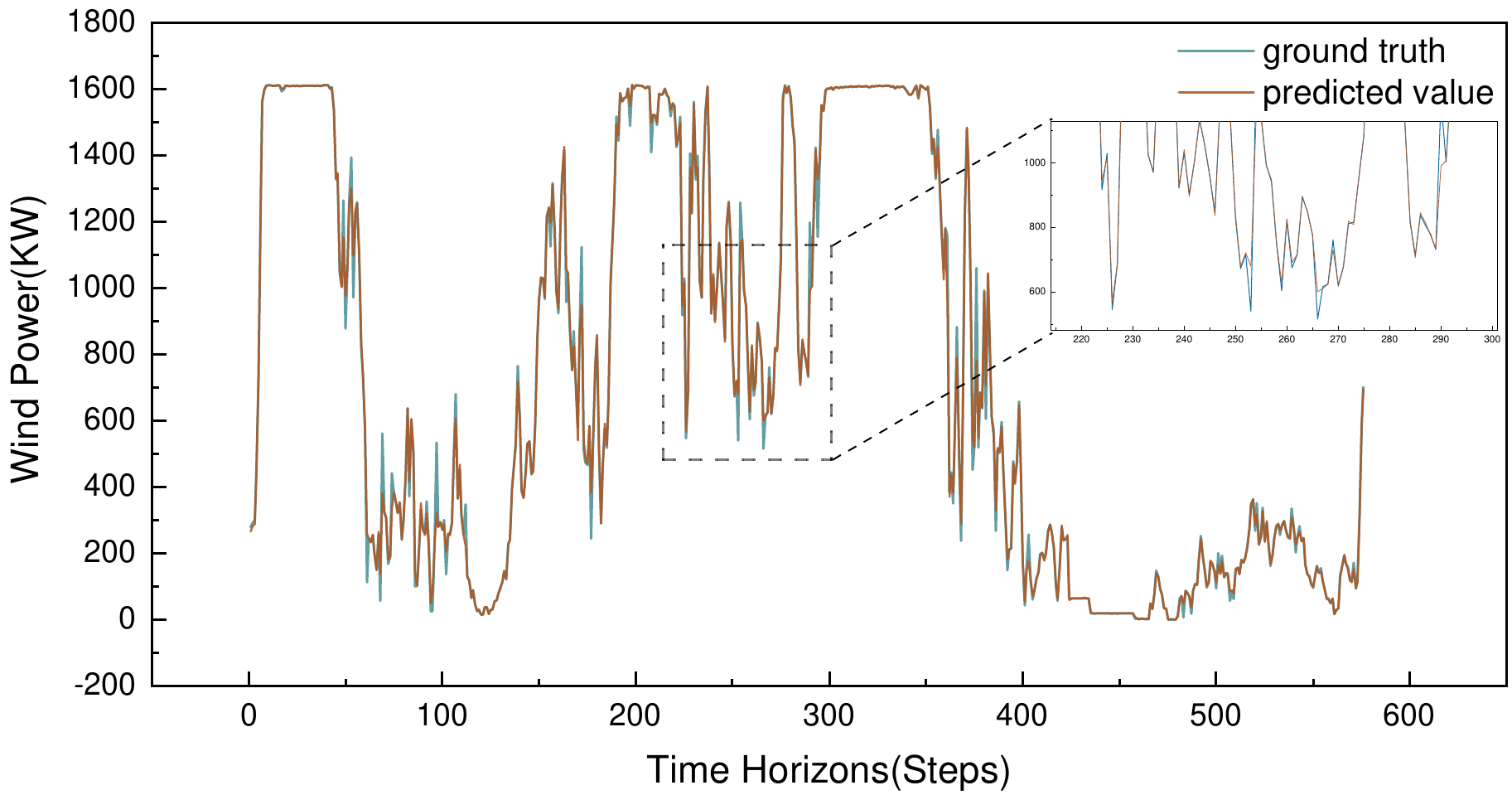}%
\label{fig3b}}
\caption{Visualization of ground truth and forecasted power values.  (a) Ground truth and forecasted power values for turbine 76 in the SDWPF dataset, spanning days 111 to 114; (b) Ground truth and forecasted power values for turbine 29 in the HHL16 dataset, from April 20th to 23rd. }
\label{fig3}
\end{figure*}
To further evaluate our proposed model, we conducted a visual inspection of the prediction results from the 2DXformer model across two datasets.  Specifically, for the SDWPF dataset, we randomly selected turbine 76 (with numbering starting from 1) and plotted its wind power output alongside the 2DXformer's predicted values from day 111 to 114, as depicted in Figure~\ref{fig3}\subref{fig3a}.  For the HHL16 dataset, we similarly selected turbine 29 and plotted its wind power output and the model's predicted values from April 20th to 23rd, as illustrated in Figure~\ref{fig3}\subref{fig3b}. 

Clearly shown in Figure~\ref{fig3}\subref{fig3a} and Figure~\ref{fig3}\subref{fig3b}, the wind power output fluctuates significantly over time and exhibits subtle periodicity, posing a significant challenge to wind power forecasting.  Volatility is greater in the HHL16 dataset compared to SDWPF, leading to poorer performance across all models on the HHL16 dataset.  However, the 2DXformer predictions on both datasets closely align with the actual values, demonstrating the accuracy of our model in forecasting. 

\section{Conclusion} \label{sec6}
In this paper, we propose the 2DXformer model to overcome the limitations present in existing wind power prediction works.  Leveraging channel independence and multi-head self-attention mechanisms, 2DXformer provides a novel approach for separately modeling endogenous and exogenous variables.  This strategy not only allows the model to account for the interrelations between variables but also minimizes unnecessary interactions between endogenous and exogenous variables.  Additionally, our solution considers spatio-temporal correlations, further enhancing the accuracy of wind power prediction. 

\section*{Acknowledgments}
This work is supported by the National Nature Science Foundation of China (Grant No.62206085), State Key Laboratory of Reliability and Intelligence of Electrical Equipment (No. EERI\_OY2022005).

\bibliography{reference}

\begin{thebibliography}{10}
\providecommand{\url}[1]{#1}
\csname url@samestyle\endcsname
\providecommand{\newblock}{\relax}
\providecommand{\bibinfo}[2]{#2}
\providecommand{\BIBentrySTDinterwordspacing}{\spaceskip=0pt\relax}
\providecommand{\BIBentryALTinterwordstretchfactor}{4}
\providecommand{\BIBentryALTinterwordspacing}{\spaceskip=\fontdimen2\font plus
\BIBentryALTinterwordstretchfactor\fontdimen3\font minus \fontdimen4\font\relax}
\providecommand{\BIBforeignlanguage}[2]{{%
\expandafter\ifx\csname l@#1\endcsname\relax
\typeout{** WARNING: IEEEtran.bst: No hyphenation pattern has been}%
\typeout{** loaded for the language `#1'. Using the pattern for}%
\typeout{** the default language instead.}%
\else
\language=\csname l@#1\endcsname
\fi
#2}}
\providecommand{\BIBdecl}{\relax}
\BIBdecl

\bibitem{wang2019approaches}
Y.~Wang, Q.~Hu, L.~Li, A.~M. Foley, and D.~Srinivasan, ``Approaches to wind power curve modeling: A review and discussion,'' \emph{Renewable and Sustainable Energy Reviews}, vol. 116, p. 109422, 2019.

\bibitem{liu2019data}
H.~Liu and C.~Chen, ``Data processing strategies in wind energy forecasting models and applications: A comprehensive review,'' \emph{Applied Energy}, vol. 249, pp. 392--408, 2019.

\bibitem{lai2023dual}
Z.~Lai and Q.~Ling, ``A dual spatio-temporal network for short-term wind power forecasting,'' \emph{Sustainable Energy Technologies and Assessments}, vol.~60, p. 103486, 2023.

\bibitem{liu2020new}
H.~Liu, C.~Yu, H.~Wu, Z.~Duan, and G.~Yan, ``A new hybrid ensemble deep reinforcement learning model for wind speed short term forecasting,'' \emph{Energy}, vol. 202, p. 117794, 2020.

\bibitem{li2022deep}
J.~Li and M.~Armandpour, ``Deep spatio-temporal wind power forecasting,'' in \emph{ICASSP 2022-2022 IEEE International Conference on Acoustics, Speech and Signal Processing (ICASSP)}.\hskip 1em plus 0.5em minus 0.4em\relax IEEE, 2022, pp. 4138--4142.

\bibitem{yu2019scene}
R.~Yu, Z.~Liu, X.~Li, W.~Lu, D.~Ma, M.~Yu, J.~Wang, and B.~Li, ``Scene learning: Deep convolutional networks for wind power prediction by embedding turbines into grid space,'' \emph{Applied energy}, vol. 238, pp. 249--257, 2019.

\bibitem{liao2022short}
W.~Liao, B.~Bak-Jensen, J.~R. Pillai, Z.~Yang, and K.~Liu, ``Short-term power prediction for renewable energy using hybrid graph convolutional network and long short-term memory approach,'' \emph{Electric Power Systems Research}, vol. 211, p. 108614, 2022.

\bibitem{ijcai2023p700}
Y.~Zhang, L.~Liu, X.~Xiong, G.~Li, G.~Wang, and L.~Lin, ``Long-term wind power forecasting with hierarchical spatial-temporal transformer,'' in \emph{Proceedings of the Thirty-Second International Joint Conference on Artificial Intelligence, {IJCAI-23}}, 2023, pp. 6308--6316.

\bibitem{wang2024timexer}
Y.~Wang, H.~Wu, J.~Dong, Y.~Liu, Y.~Qiu, H.~Zhang, J.~Wang, and M.~Long, ``Timexer: Empowering transformers for time series forecasting with exogenous variables,'' \emph{arXiv preprint arXiv:2402.19072}, 2024.

\bibitem{liu2024itransformer}
Y.~Liu, T.~Hu, H.~Zhang, H.~Wu, S.~Wang, L.~Ma, and M.~Long, ``itransformer: Inverted transformers are effective for time series forecasting,'' in \emph{The Twelfth International Conference on Learning Representations}, 2024.

\bibitem{bartholomew1971time}
D.~J. Bartholomew, ``Time series analysis forecasting and control.'' 1971.

\bibitem{williams2001multivariate}
B.~M. Williams, ``Multivariate vehicular traffic flow prediction: Evaluation of arimax modeling,'' \emph{Transportation Research Record}, vol. 1776, no.~1, pp. 194--200, 2001.

\bibitem{vagropoulos2016comparison}
S.~I. Vagropoulos, G.~Chouliaras, E.~G. Kardakos, C.~K. Simoglou, and A.~G. Bakirtzis, ``Comparison of sarimax, sarima, modified sarima and ann-based models for short-term pv generation forecasting,'' in \emph{2016 IEEE international energy conference (ENERGYCON)}.\hskip 1em plus 0.5em minus 0.4em\relax IEEE, 2016, pp. 1--6.

\bibitem{amjady2011wind}
N.~Amjady, F.~Keynia, and H.~Zareipour, ``Wind power prediction by a new forecast engine composed of modified hybrid neural network and enhanced particle swarm optimization,'' \emph{IEEE transactions on sustainable energy}, vol.~2, no.~3, pp. 265--276, 2011.

\bibitem{liang2016short}
Z.~Liang, J.~Liang, C.~Wang, X.~Dong, and X.~Miao, ``Short-term wind power combined forecasting based on error forecast correction,'' \emph{Energy conversion and management}, vol. 119, pp. 215--226, 2016.

\bibitem{kong2015wind}
X.~Kong, X.~Liu, R.~Shi, and K.~Y. Lee, ``Wind speed prediction using reduced support vector machines with feature selection,'' \emph{Neurocomputing}, vol. 169, pp. 449--456, 2015.

\bibitem{das2023longterm}
A.~Das, W.~Kong, A.~Leach, S.~K. Mathur, R.~Sen, and R.~Yu, ``Long-term forecasting with ti{DE}: Time-series dense encoder,'' \emph{Transactions on Machine Learning Research}, 2023.

\bibitem{girshick2015fast}
R.~Girshick, ``Fast r-cnn,'' in \emph{Proceedings of the IEEE international conference on computer vision}, 2015, pp. 1440--1448.

\bibitem{zhou2022sdwpf}
J.~Zhou, X.~Lu, Y.~Xiao, J.~Su, J.~Lyu, Y.~Ma, and D.~Dou, ``Sdwpf: A dataset for spatial dynamic wind power forecasting challenge at kdd cup 2022,'' \emph{arXiv preprint arXiv:2208.04360}, 2022.

\bibitem{cho2014learning}
K.~Cho, B.~Van~Merri{\"e}nboer, C.~Gulcehre, D.~Bahdanau, F.~Bougares, H.~Schwenk, and Y.~Bengio, ``Learning phrase representations using rnn encoder-decoder for statistical machine translation,'' \emph{arXiv preprint arXiv:1406.1078}, 2014.

\bibitem{vaswani2017attention}
A.~Vaswani, N.~Shazeer, N.~Parmar, J.~Uszkoreit, L.~Jones, A.~N. Gomez, {\L}.~Kaiser, and I.~Polosukhin, ``Attention is all you need,'' \emph{Advances in neural information processing systems}, vol.~30, 2017.

\bibitem{bai2020adaptive}
L.~Bai, L.~Yao, C.~Li, X.~Wang, and C.~Wang, ``Adaptive graph convolutional recurrent network for traffic forecasting,'' \emph{Advances in neural information processing systems}, vol.~33, pp. 17\,804--17\,815, 2020.

\bibitem{jiang2023spatio}
R.~Jiang, Z.~Wang, J.~Yong, P.~Jeph, Q.~Chen, Y.~Kobayashi, X.~Song, S.~Fukushima, and T.~Suzumura, ``Spatio-temporal meta-graph learning for traffic forecasting,'' in \emph{Proceedings of the AAAI conference on artificial intelligence}, vol.~37, no.~7, 2023, pp. 8078--8086.

\bibitem{liu2023spatio}
H.~Liu, Z.~Dong, R.~Jiang, J.~Deng, J.~Deng, Q.~Chen, and X.~Song, ``Spatio-temporal adaptive embedding makes vanilla transformer sota for traffic forecasting,'' in \emph{Proceedings of the 32nd ACM international conference on information and knowledge management}, 2023, pp. 4125--4129.

\end{thebibliography}
\end{document}